\documentclass[final,5p,times,twocolumn]{elsarticle}

\usepackage[numbers]{natbib}

\usepackage{graphics} 
\usepackage{amsmath} 
\usepackage{amssymb}  

\usepackage{amsmath,amsfonts}
\usepackage{algorithmic}
\usepackage{algorithm}
\usepackage{array}
\usepackage{textcomp}
\usepackage{stfloats}
\usepackage{url}
\usepackage{verbatim}
\usepackage{graphicx}
\usepackage{cite}

\usepackage{booktabs}

\usepackage{color} 
\floatname{}{algorithm}

\usepackage{mathtools}

\usepackage{multirow}
\usepackage{tabu}
\usepackage[pagebackref=true,breaklinks=true,colorlinks,bookmarks=false]{hyperref}

\usepackage{subfigure}
\usepackage{tikz} 
\usepackage{overpic}
\usepackage{balance}

\usepackage{xcolor}

\journal{Image and Vision Computing}

\begin{document}

\begin{frontmatter}



\title{Density-aware Global-Local Attention Network for Point Cloud Segmentation}


\author[add1,add2]{Chade Li} 
\ead{lichade2021@ia.ac.cn}

\author[add1]{Pengju Zhang\corref{cor1}}
\ead{pengju.zhang@ia.ac.cn}

\author[add1]{Jiaming Zhang}
\ead{jm-zhang21@mails.tsinghua.edu.cn}

\author[add1,add2]{Yihong Wu\corref{cor1}}
\ead{yhwu@nlpr.ia.ac.cn}
\cortext[cor1]{Corresponding authors}

\address[add1]{State Key Laboratory of Multimodal Artificial Intelligence Systems, Institute of Automation, Chinese Academy of Sciences, Beijing 100190, China}
\address[add2]{School of Artificial Intelligence, University of Chinese Academy of Sciences, Beijing 100049, China}

\begin{abstract}
3D point cloud segmentation has a wide range of applications in areas such as autonomous driving, augmented reality, virtual reality and digital twins. The point cloud data collected in real scenes often contain small objects and categories with small sample sizes, which are difficult to handle by existing networks. In this regard, we propose a point cloud segmentation network that fuses local attention based on density perception with global attention. The core idea is to increase the effective receptive field of each point while reducing the loss of information about small objects in dense areas. Specifically, we divide different sized windows for local areas with different densities to compute attention within the window. Furthermore, we consider each local area as an independent token for the global attention of the entire input. A category-response loss is also proposed to balance the processing of different categories and sizes of objects. In particular, we set up an additional fully connected layer in the middle of the network for prediction of the presence of object categories, and construct a binary cross-entropy loss to respond to the presence of categories in the scene. In experiments, our method achieves competitive results in semantic segmentation and part segmentation tasks on several publicly available datasets. Experiments on point cloud data obtained from complex real-world scenes filled with tiny objects also validate the strong segmentation capability of our method for small objects as well as small sample categories.	
\end{abstract}


\end{frontmatter}



\section{Introduction}

As one of the important research directions in the field of 3D computer vision, 3D point cloud semantic segmentation has a wide range of applications in autonomous driving, AR/VR, and digital twins, etc. With the proposal of Transformer \citep{12-attention} network, which is based on the attention mechanism, the point cloud segmentation network has been further improved. According to the processing scale of the attention mechanism, 3D Transformers can be classified into two main categories: global Transformers and local Transformers. The former directly performs attention computation on all input points for global feature extraction; the latter first divides the point cloud into multiple windows and performs attention computation within the windows to extract local features.

Global Transformer point cloud networks (PCT \citep{13-pct}, 3CROSSNet \citep{36-3crossnet}, Point-BERT \citep{37-pointbert}, etc.) tend to be applied only to object-level tasks and cannot effectively handle scene-level point cloud inputs due to the large amount of computation required by itself. Moreover, this type of method requires sampling when processing point clouds, which tends to ignore categories with fewer samples or small objects, and focuses more on the main part of the point cloud, thus tends to be more suitable for tasks such as classification.

\begin{figure}[t]
\centering
\includegraphics[width=0.45\textwidth]{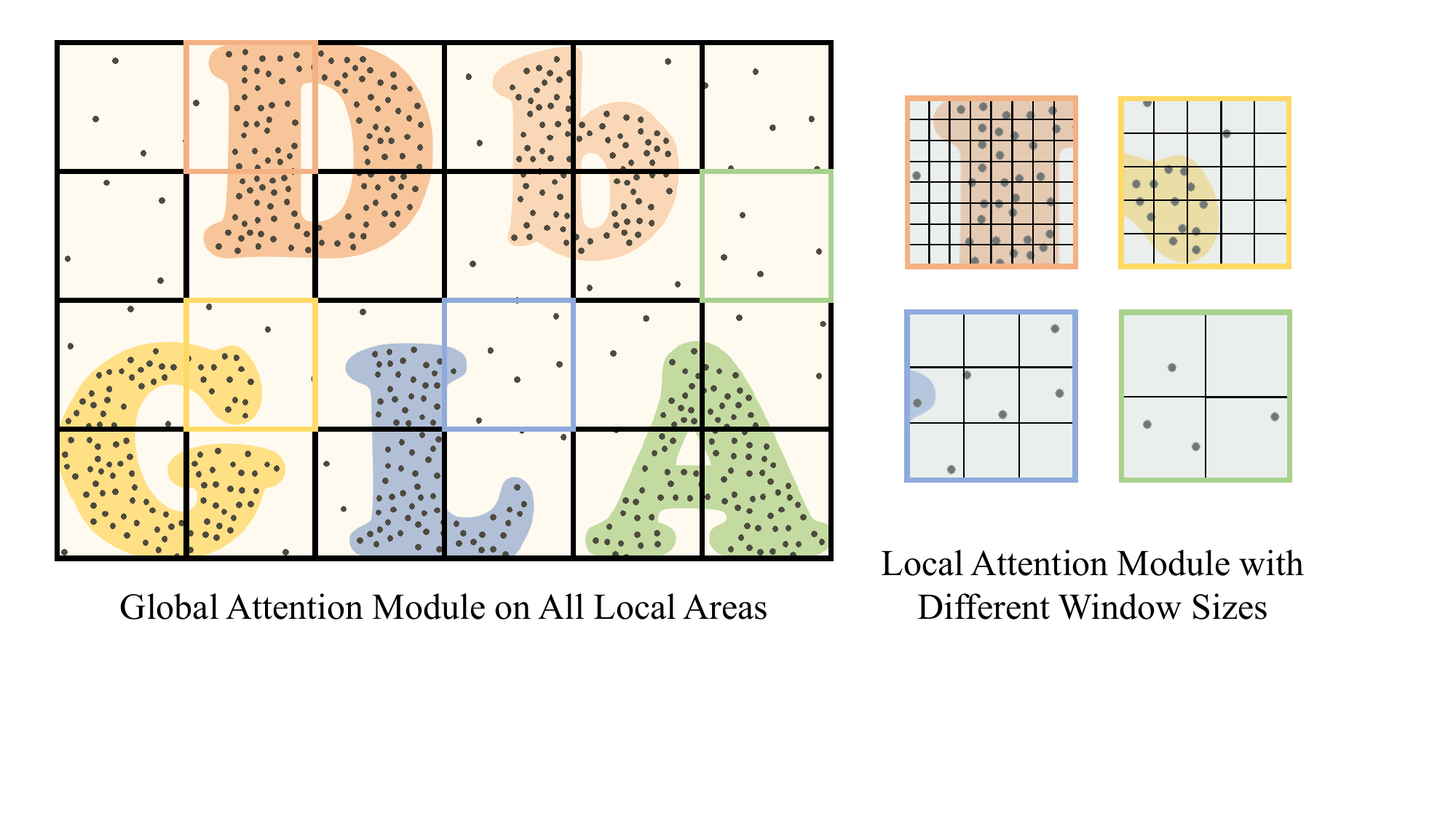}
\caption{Global attention carried out between local areas and local attention with different window sizes within local areas of different densities.}
\label{fig1}
\end{figure}

Existing local Transformer networks (Point Transformer \citep{14-pt}, Stratified Transformer \citep{16-st}, PPT-Net \citep{35-pptnet}) tend to divide the whole point cloud into windows of uniform size and perform the local attention. Due to the different distances of objects from the acquisition device in space and the difference in the amount of objects in different areas, the densities of different areas in point cloud tend to be unbalanced. And these networks mentioned above need to sample the point cloud before performing local attention, so some of the point in each local area would be randomly discarded due to the limited computational resources. As a result, the networks tend to learn poorly for some of the categories, especially the small object categories distributed in the dense area of the point cloud, thus affecting the recognition accuracy of the corresponding categories. In addition, the information interaction range of local attention networks is often limited within a single window, or the information interaction range is further improved by moving window operations such as Stratified Transformer \citep{16-st}, but the existing operations like shifting window are limited in terms of the enhancement of the effective receptive field.

Aiming at the above problems, we propose a 3D point cloud segmentation method that fuses global attention with density-aware local attention. Unlike other methods that fix the local window size for attention of each network layer, the entire input point cloud is first adaptively divided into a controllable number of equal-sized local areas according to its true dimensional size. Based on the number of 3D points in each local area, as shown in Figure \ref{fig1}, the local density is calculated, and all local areas are divided into several discrete parts according to different densities. Then different local window sizes are determined for local areas belonging to different discrete parts in the local attention module, in order to density-aware processing as shown in Figure \ref{fig2}. At the same time, each local area is regarded as an independent whole, and an additional module is set up to calculate the global attention between different local areas. The global and local attention are computed in parallel, and the fusion between local and global attention is performed in the upsampling stage of the network, so that each 3D point can get the global-local feature update. In addition, a 3D category-response loss function is proposed, which allows the network to pay attention to categories with different sample sizes and to objects of different sizes more balanced among them.

Following the introduction, Section 2 (Related Work) reviews existing literature on 3D point cloud segmentation and Transformer-based networks. Section 3 (Methodology) details the proposed approach, introducing the Density-based Global-Local Attention mechanism (with subsections on local and global feature updates) and the novel loss function. Section 4 (Experiments) comprehensively evaluates the method, describing the experimental settings, and presenting results on indoor scene segmentation, outdoor scene segmentation, part segmentation, and a series of ablation studies. Finally, the paper concludes with Section 5 (Conclusions), summarizing the findings, contributions, limitation and future work.

The main contributions of this paper are: 
\begin{itemize}
    \item Our work pioneers the integration of local, density-aware attention and global attention. It dynamically reassigns feature aggregation boundaries via DBSCAN-based density partitioning, replacing static geometric partitioning.
    \item The proposed attention mechanism divides point clouds into discrete density parts, each of which is assigned a local attention window. This achieves higher accuracy by focusing computational resources intensively on detailed information in dense regions.
    \item The introduction of category-response loss (CR-loss) establishes an intermediate supervisory mechanism that enforces category-specific feature separation at the early stages of the network, collectively forming a distinct point cloud segmentation framework architecturally.
    \item Exhaustive comparison experiments on commonly used public datasets as well as self-acquisition datasets and ablation experiments show that our proposed method has excellent segmentation results, validating the effectiveness and robustness of our work.
\end{itemize}

\section{Related Work}
\subsection{3D Point Cloud Segmentation}
The goal of the 3D point cloud segmentation is to process the point cloud input to obtain the category labels corresponding to each 3D point. Commonly used 3D point cloud segmentation methods can be classified into the following categories based on the principles of the network and the data format of the input: voxel-based methods \citep{4-subsc, 5-4dstc}, projection-based methods \citep{1-deep3ds,2-squeezeseg,3-polarnet}, MLP-based methods \citep{8-pointnet,9-pointnet++,10-deepgcns,11-randlanet,28-thickseg}, pointwise convolution-based methods \citep{6-pointcnn,7-kpconv,26-stpca,30-hcfs3d}, graph convolution-based methods \citep{29-mdcgcns, 31-dgcnn}, methods combining image processing \citep{24-3dmv,25-xmuda}, and attention-based methods \citep{13-pct,36-3crossnet,37-pointbert,14-pt,16-st,35-pptnet,15-fpt,17-ptv2,18-superpt,19-spotr,20-supercluster,26-stpca,27-iamtm}. 

SSCNs \citep{4-subsc} introduces Submanifold Sparse Convolutional Networks, which efficiently process sparse 3D point clouds by restricting convolutions to non-empty voxels for semantic segmentation. MinkowskiNet \citep{5-4dstc} proposes Minkowski Convolutional Neural Networks using 4D spatio-temporal sparse tensors to achieve large-scale spatiotemporal point cloud segmentation. These voxel-based methods discretise the 3D space into regular grids to enable 3D convolutions, but often struggle with trade-offs between resolution and efficiency. F. J. Lawin et al. \citep{1-deep3ds} design a deep projective 3D segmentation framework by projecting point clouds to 2D multi-views and fusing features to enhance semantic segmentation accuracy. Sueezeseg \citep{2-squeezeseg} combines CNNs with CRF to project 3D LiDAR point clouds onto 2D spherical grids for real-time road-object segmentation. PolarNet \citep{3-polarnet} improves polar grid representation using annular partitioning to enhance LiDAR point cloud segmentation in bird's-eye view. These projection-based methods transform 3D points into 2D representations, leveraging mature 2D CNN architectures for feature learning. MLP-based methods, such as PointNet \citep{8-pointnet} and its successors PointNet++ \citep{9-pointnet++}, DeepGCNs \citep{10-deepgcns}, RandLA-Net \citep{11-randlanet}, ThickSeg \citep{28-thickseg}, operate directly on raw point clouds, introducing innovative mechanisms to handle unordered data. Recent advances in point-wise convolution and attention mechanisms have further enhanced the ability to capture local and global geometric relationships.

\begin{figure*}[t]
\centering
\includegraphics[width=0.85\textwidth]{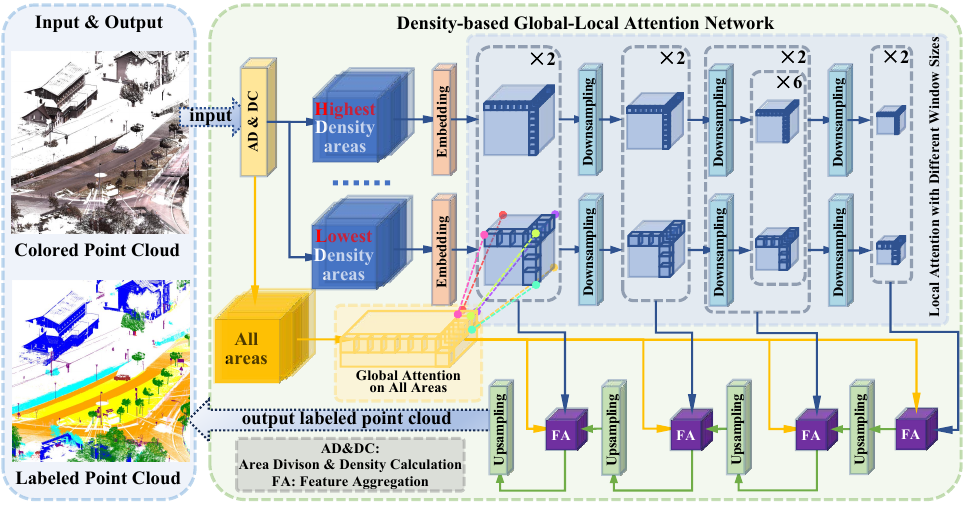}  
\caption{Overview of proposed methods (taking outdoor scenarios as an example).}
\label{fig2}
\end{figure*}

\begin{figure}[t]
\centering
\includegraphics[width=0.4\textwidth]{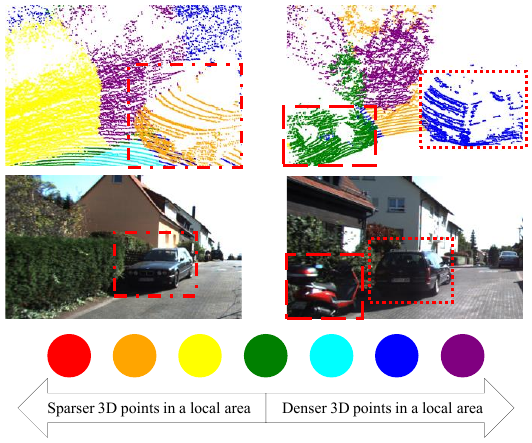}
\caption{Visualisation diagram of local area density division of point cloud (the visualisation of point cloud file density division in SemanticKITTI \citep{38-semantickitti}.}
\label{fig3}
\end{figure}


\subsection{Network based on Transformer}
Since Transformer \citep{12-attention} networks have a substantial increase in processing performance compared to CNN networks in their original field of Natural Language Processing (NLP), many articles in the vision field have attempted to use the Transformer-based network. In the field of 3D vision processing, the Transformer architecture is applied to similar tasks in 3D.

Most of the current point cloud transformer networks \citep{14-pt,16-st,15-fpt,17-ptv2} only consider local features without considering long-range dependencies, and are unable to better extract complete global context information. Point Transformer \citep{14-pt} innovatively applies transformer-based self-attention mechanisms to 3D point clouds, adaptively modeling contextual relationships between points with geometrically weighted attention to capture local structures and enhance feature representation. Stratified Transformer \citep{16-st} proposes a hierarchical architecture that processes 3D point clouds at multiple scales with stratified self-attention, capturing both fine-grained details and global contextual information for improved semantic segmentation. Fast Point Transformer \citep{15-fpt} introduces an efficient framework that accelerates self-attention computation for point clouds via a hierarchical grouping strategy and approximate attention, enabling real-time processing while maintaining segmentation accuracy. Point Transformer V2 \citep{17-ptv2} advances the framework with grouped vector attention, which models point features as vector groups to capture directional relationships, and partition-based pooling, which enhances spatial context aggregation for more discriminative 3D feature learning. The above attention methods based on the transformer network architecture, although it expands the range of sensory field and information interaction to a certain extent, part of the long-range information is still not considered, and the window division also affects the performance. In real scenes such as oil field production, there are a large number of empty or weakly textured areas, and at the same time, there are small valves, railings, pipelines and other slender objects, and it is difficult for existing methods to perform point cloud segmentation better. Therefore, the proposed method is designed in terms of local window segmentation strategy, global feature update path, and loss function to further solve the problems existing in the current network.

\section{Methodology}
The framework of the point cloud segmentation network fusing global attention and density-aware local attention proposed in this paper is shown in Figure \ref{fig2}. Firstly, in the point cloud local area division and density calculation module, the input point cloud is divided into local areas, the point cloud density is calculated for each local area and the corresponding local density value is assigned, then all the local areas are divided into multiple discrete parts according to the specific density, and the proposed Density-based Global-Local Attention module (DbGLA) is used for feature processing of point cloud data.

\subsection{Density-based Global-Local Attention}

The proposed DbGLA module is mainly divided into two main parts, namely the local feature update module and the global feature update module: the former determines the window sizes inside the local areas based on their densities and computes the local attention between the internal windows within each local area; the latter treats each local area as an independent whole, and computes the global attention between different local areas of the whole scene.

\begin{figure*}[t]
\centering
\includegraphics[width=0.85\textwidth]{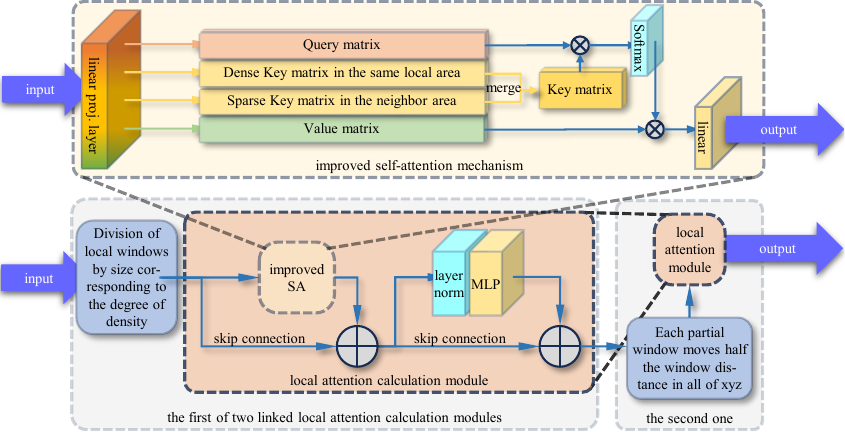}
\caption{A schematic diagram of two connected local attention calculation modules.}
\label{fig4}
\end{figure*}


\subsubsection{Local Feature Updates}
Conventional Transformer-based 3D point cloud semantic segmentation networks \citep{13-pct,14-pt,16-st,15-fpt,17-ptv2}, due to the large computational resources they require, need to first divide the input point cloud into grids of uniform size and sample the panels when there are too many points in them for local feature updating. However, the density of the point cloud data varies in different local areas, and setting a uniform grid size will cause the dense area of the overall point cloud to discard more 3D points compared to the sparse area when sampling, resulting in the loss of too many 3D points of smaller objects in the dense area. In this regard, the overall point cloud is first divided into multiple local areas of the same size. Then we set the number of local areas to be divided, and calculate the size of the local areas according to the number of areas and the real space occupied by the input point cloud. 

Then we use the idea of DBSCAN\citep{32-dbscan} to partition different local regions of the input point cloud according to different densities. The DBSCAN algorithm, a density-based clustering method, operates by first selecting an unvisited point in the dataset and identifying its neighboring points within a specified radius, $\epsilon$. If the number of neighboring points is greater than or equal to a predefined threshold, the point is marked as a core point and a new cluster is initiated. Then, the algorithm recursively expands the cluster by adding all density-reachable points to the cluster, which are those within the $\epsilon$ radius of the core point. This process continues, incorporating neighboring points and forming a dense cluster. If a point does not meet the density criteria, it is labeled as noise. The algorithm iterates through the dataset, repeating this process for each unvisited core point until all points have been visited and assigned to a cluster or labeled as noise. This approach allows DBSCAN to identify clusters of arbitrary shapes and effectively handle outliers. After partition different local areas of the input point cloud according to different densities, all areas are sorted from sparse to dense density, divided into multiple discrete parts of density. We then assign a density value to the corresponding local area and all 3D points contained within it, with the sparsest area having the smallest density value and vice versa.

We set the same window size for local areas belonging to the same discrete part of the point cloud density, setting larger windows for areas with smaller density values and vice versa. As shown in Fig. \ref{fig3} is the visualisation of the outdoor scene point cloud data processed using the proposed local area density division module. The different colours represent the different density degree of the corresponding local areas of the point cloud, and it can be seen that the local areas containing special categories (e.g., car, motorcycle, etc., which are boxed by red) are distinctly different from the density degree assigned to the neighbouring areas.

In order to increase the effective receptive field, we build cross-window information interaction. The local attention module designed is shown in Figure \ref{fig4}.

After identifying and dividing the small windows within each local area, we introduce the density-aware local attention. For Query points within a specific window, the selection of Key points consists of two main parts, which are densely sampling points within the same window and sparsely sampling points within neighbouring windows as well as other windows with the same degree of density. In addition, the multi-head self-attention is independently used within each small window. Given that the number of heads for calculating the attention is $n_h$ and the dimension of each attention head is $n_d$, and assuming that the number of 3D points within the $i-$th small window is $n_i$. Then the feature dimensions of all the input belonging to the $i-$th small window can be obtained as $x \in \mathbb{R} ^ {( n_d \times n_h ) \times n_i}$. As in Equation \ref{eq1} to Equation \ref{eq6} is the computational procedure for calculating the local multi-head self-attention for the corresponding window:

\begin{equation}
Q,K,V=Linear_{q,k,v}^{\mathbb{R}^{(n_d \times n_h) \times n_i} \rightarrow \mathbb{R}^{n_d \times n_h \times n_i}} (x),
\label{eq1}
\end{equation}



\begin{equation}
attn_{i,j,h}=softmax(Q_{i,h} \cdot K_{j,h}),
\label{eq4}
\end{equation}

\begin{equation}
out_{i,h}=\sum_{j=1}^{n_i} (attn_{i,j,h} \times V_{j,h}),
\label{eq5}
\end{equation}

\begin{equation}
o=Linear^{\mathbb{R}^{n_d \times n_h \times n_i} \rightarrow \mathbb{R}^{(n_d \times n_h) \times n_i}} (out),
\label{eq6}
\end{equation}

\noindent where $Linear$ represents the linear layer operation and its superscripts represent the input and output dimensions of this linear layer. The $attn_{i,j,h}$ is the attention value computed by dot product of the vectors corresponding to the $i$th Query point and the $j$th Key point of the $h$th attention head in the window. The $out$ is then the aggregated features obtained by the computation. Finally, the features are projected back to the dimension consistent with the input features through an additional linear layer to obtain the final output feature $o$.

Moreover, we choose to adaptively compute and adjust the local area size, so that the total number of local areas for different inputs can be similar, which facilitates the subsequent computation of the global attention. Also the above operation avoids excessive computation due to too many local areas in the case of a large input point cloud.

\subsubsection{Global Feature Updates}
For the global feature updates, we regard each local area as a whole that aggregates the position and feature information of the corresponding area to calculate the attention map between different local areas. Before calculating the global attention, the information for local area needs to be aggregated, and this aggregation process is mainly divided into two steps, the first step is the aggregation of 3D spatial location information, as shown in Equation \ref{eq7}; the second step is the aggregation of feature information of the local area, as shown in Equation \ref{eq8}:

\begin{equation}
p_i^{'}=\frac{\sum_{p_j \in \mathcal{P}_i}{p_j}}{N_i},
\label{eq7}
\end{equation}

\begin{equation}
f_i^{'}=\sum_{p_j \in \mathcal{P}_i}{(f_{p_j} \bigoplus e_{p_j})},~ e_{p_j}=Linear_{enc}^{\mathbb{R}^3 \rightarrow \mathbb{R}^{enc}} (p_j - p_i^{'}),
\label{eq8}
\end{equation}

\noindent where $p_i^{'}$ represents the spatial coordinates after local area aggregation. $f_i^{'}$ is the aggregated feature of the corresponding area. $\bigoplus$ represents the vector concatenation operation. $p_j \in \mathcal{P}_i$ denotes that the 3D point $p_j$ belongs to the $i$th local area $P_i$, $N_i$ denotes the number of 3D points in the $i$th local area, $p_j$ and $f_{p_j}$ stand for the spatial coordinates of the corresponding 3D point and the input features of the corresponding point, $e_{p_j}$ represents the relative position encoding of the 3D point $p _j$ and the relative position encoding of the centre of gravity $p_i^{'}$ of the local area to which it belongs.

In addition, after completing the global attention, the updated information needs to be assigned to 3D points, and again this process can be divided into two major parts, which are the assignment of spatial coordinates of the 3D points $p_j^{out}=p_j$, and the process of assigning global features of the 3D points as shown by Equation \ref{eq10}:

\begin{equation}
f_{p_j}^{out}=Linear^{\mathbb{R}^{fea+enc} \rightarrow \mathbb{R}^{fea}} (f_i^{'}),~ p_j \in \mathcal{P}_i,
\label{eq10}
\end{equation}

\noindent where $p_j^{out}$ and $f_{p_j}^{out}$ represent the spatial coordinates and feature information of the corresponding 3D points after the global feature update, respectively.

\subsection{Loss Function}
When training the network, the loss function applied in this paper is mainly divided into two parts, which are the weighted cross-entropy loss function that adaptively determines the weights of the categories according to different datasets, and the category-response loss function oriented to 3D point cloud processing, and fuses the two kinds of loss functions according to the weights given by the design. When using the ordinary cross-entropy loss function for network model training, if the sample number of individual categories is too large, the loss function will make the network model pay more attention to the category, thus affecting the processing accuracy of the network model for the category with too few samples. In this regard, this paper adaptively calculates the number of 3D points in each category of samples in the training set for different experimental datasets, and calculates the weights of different categories in the calculation of the cross-entropy loss function, and the cross-entropy loss function after adding the weights is shown in Equation \ref{eq11}:

\begin{equation}
\mathcal{L}_{WCE}=-\frac{1}{N} \sum_{i=1}^N \sum_{k=0}^{C-1} w_k l_{ik} \log(p_{ik}),
\label{eq11}
\end{equation}

\begin{table*}[!t]
\caption{Quantitative comparison experiment and ablation experiment results of semantic segmentation of indoor point cloud dataset S3DIS \citep{21-s3dis} Area 5.}
\label{tab1}
\centering
{\scriptsize
\begin{tabular}{l|ccc|ccccccccccccc}
\toprule
  Method & OA & mAcc & mIoU & ceiling & floor & wall & beam & column & window & door & table & chair & sofa & bookcase & board & clutter\\\hline
  PointNet \citep{8-pointnet} & - & 49.0 & 41.1 & 88.8 & 97.3 & 69.8 & \underline{0.1} & 3.9 & 46.3 & 10.8 & 59.0 & 52.6 & 5.9 & 40.3 & 26.4 & 33.2\\
  KPConv \citep{7-kpconv} & - & 72.8 & 67.1 & 92.8 & 97.3 & 82.4 & 0.0 & 23.9 & 58.0 & 69.0 & 91.0 & 81.5 & 75.3 & 75.4 & 66.7 & 58.9\\
  PCT \citep{13-pct} & - & 67.7 & 61.3 & 92.5 & 98.4 & 80.6 & 0.0 & 19.4 & \underline{61.6} & 48.0 & 85.2 & 76.6 & 67.7 & 46.2 & 67.9 & 52.3\\
  ST \citep{16-st} & 91.5 & \underline{78.1} & \underline{72.0} & \textbf{96.2} & \textbf{98.7} & 85.6 & 0.0 & 46.1 & 60.0 & \textbf{76.8} & \textbf{92.6} & 84.5 & 77.8 & 75.2 & \textbf{78.1} & \underline{64.0}\\
  SpT \citep{18-superpt} & 89.5 & 77.3 & 68.9 & 92.6 & 97.7 & 83.5 & \textbf{0.2} & 42.0 & 60.6 & 67.1 & 88.8 & 81.0 & 73.2 & \textbf{86.0} & 63.1 & 60.0\\
  SPoTr \citep{19-spotr} & 90.7 & 76.4 & 70.8 & - & - & - & - & - & - & - & - & - & - & - & - & -\\\hline
  w/o DbGLA & \underline{91.6} & 76.8 & 71.0 & 95.2 & \underline{98.6} & \underline{87.0} & 0.0 & \underline{46.6} & 61.1 & 74.7 & 89.9 & \textbf{85.3} & \textbf{78.8} & 68.7 & 73.2 & 63.9\\
  w/ DbGLA & \textbf{92.1} & \textbf{79.0} & \textbf{72.5} & \underline{96.1} & 98.5 & \textbf{87.9} & 0.0 & \textbf{48.7} & \textbf{61.7} & \underline{76.5} & \underline{92.3} & \underline{85.2} & \underline{78.5} & \underline{77.3} & \underline{75.6} & \textbf{64.1}\\
\bottomrule
\end{tabular}}
\end{table*}

\noindent where $C$ represents the total number of categories. $N$ represents the total number of points. $w_k$ represents the adaptive weights of the $k$th category in calculating the cross-entropy loss, which is inversely proportional to the ratio of the number of samples of the corresponding category points to the total number of points extracted from the partially randomly selected scenarios multiple times. $l_{ik}$ represents the labelling of whether the $i-$th point belongs to the $k-$th category or not, when the $i-$th point belongs to the $k-$th category, the corresponding $l_{ik}$ value is $1$, otherwise it is $0$. The $p_{ik}$ represents the probability that the network predicts the $k-$th category for the $i-$th point.

In addition to this, a novel 3D category-response loss function (CR-Loss) is proposed to enhance the network's learning for semantic contexts as shown in Equation \ref{eq12}:

\begin{equation}
\mathcal{L}_{3DCR}=-\frac{1}{N} \sum_{k=0}^{C-1} \sum_{i=1}^N (s_{ik}^l \log(s_{ik}^p) + (1 - s_{ik}^l) \log(1 - s_{ik}^p)),
\label{eq12}
\end{equation}

\noindent where $s_{ik}^l$ represents the binary classification labels of the $i-$th point against the $k-$th category, if the $i-$th point belongs to the $k-$th category, the corresponding $s_{ik}^l$ value is $1$, otherwise it is $0$. The $s_{ik}^p$ represents the binary prediction result for the $i-$th point against the $k-$th category.

Unlike pixel-level loss, CR-Loss contributes equally to both large and small targets, and tends to improve substantially in practice for segmentation processing performance for small targets. In the standard training process for 3D semantic segmentation, the network usually using cross-entropy loss computed between the prediction and ground truth for each point, and they may have difficulty understanding the context without global information. In order to standardise the training of the context encoding module, we introduce category response loss using the category information embedded in the scene during processing.

An additional fully-connected layer with a sigmoid activation function is constructed in addition to the coding layer of the proposed network in order to make individual predictions about the presence of object categories in the scene and learnt through binary cross-entropy loss, a setup that forces the network to understand the global semantic information at a very small additional computational cost. In subsequent experiments, we will see that the inclusion of 3D CR-Loss significantly improves the segmentation of small targets.

From combining the two loss functions, the final loss $\mathcal{L}_{all}$ are the sum of $\mathcal{L}_{WCE}$ and $\mathcal{L}_{3DCR}$ using a hyperparameter $\lambda$ as shown in Equation \ref{eq13}:

\begin{equation}
\mathcal{L}_{all}=\lambda \mathcal{L}_{WCE} + (1 - \lambda) \mathcal{L}_{3DCR}.
\label{eq13}
\end{equation}

\section{Experiments}
In order to verify the effectiveness of the proposed method, we choose compare it with existing representative methods on several commonly used indoor and outdoor 3D point cloud datasets. The proposed network is also applied to the part segmentation task for the corresponding experiments. In addition, we collect point cloud data from oil field production scenarios and perform segmentation experiments on these outdoor real field point clouds to verify the generality of the proposed method. The designed ablation experiments evaluate the effectiveness of the DbGLA module with the proposed CR-Loss.

\subsection{Experimental settings}
\subsubsection{Comparison Methods}
In this paper, the proposed methods are compared with existing representative methods applied to the task of semantic segmentation of 3D point clouds, respectively. Specifically, we choose MLP-based methods (PointNet \citep{8-pointnet}, PointNet++ \citep{9-pointnet++}, RandLA-Net \citep{11-randlanet}), point convolution methods (KPConv \citep{7-kpconv}, OA-CNNs \citep{42-oacnns}, MinkUNet \citep{48-minkunet}, SPVNAS \citep{49-spvnas}, Cylinder3D \citep{50-cylinder}), methods combining image processing (WaffleIron \citep{40-waffleiron}, 2DPASS \citep{41-2dpass}, FRNet \citep{46-frnet}) and other Transformer network-based methods for attention mechanisms, which include the latest state-of-the-art methods that have been published in recent years: Point Cloud Transformer \citep{13-pct}, Point Transformer \citep{14-pt}, Stratified Transformer \citep{16-st}, Superpoint Transformer \citep{18-superpt}, SPoTr \citep{19-spotr}, SuperCluster \citep{20-supercluster}, Point Transformer V2 \citep{17-ptv2}, OctFormer \citep{34-octformer}, SphereFormer \citep{39-sphereformer}, Point Transformer V3 \citep{44-ptv3}.

\subsubsection{Datasets}
Following the practice of new point cloud segmentation models in recent years, we conducted our semantic segmentation experiments on six popular benchmarks, including the indoor scene datasets S3DIS \citep{21-s3dis}, ScanNet \citep{33-scannet}, as well as the outdoor scene datasets Semantic3D \citep{22-semantic3d}, SemanticKITTI \citep{38-semantickitti}, nuScenes \citep{45-nuscenes} and Waymo Open \citep{47-waymo}. S3DIS \citep{21-s3dis} contains 271 indoor rooms across 6 large-scale areas, providing point cloud data with 13 semantic class annotations (e.g., furniture, walls), tailored for 3D semantic and instance segmentation in indoor scenes. ScanNet \citep{33-scannet} includes over 1500 indoor scans with 2.5 million RGB-D views, offering 3D camera poses, surface reconstructions, and instance-level semantic labels, supporting point cloud segmentation and scene understanding tasks. Semantic3D \citep{22-semantic3d} comprises over 3 billion outdoor points across urban streets, churches, etc., labeled with 8 semantic classes (e.g., buildings, vegetation), designed for large-scale 3D semantic segmentation in complex outdoor environments. SemanticKITTI \citep{38-semantickitti} extends KITTI with 22 sequences of LiDAR point clouds, distinguishing dynamic/static objects via dense annotations for 22 classes, enabling temporal point cloud segmentation in autonomous driving scenarios. NuScenes \citep{45-nuscenes} captures multi-modal data (LiDAR, camera, radar) across 2 cities (Boston, Singapore), with its LiDAR Seg subset providing 32-class annotations for 23 object categories, supporting point cloud segmentation in complex traffic scenes. Waymo Open Dataset \citep{47-waymo} integrates LiDAR, camera, and radar data across 5000+ driving segments (including 2025 additions), focusing on long-tail outdoor scenarios (e.g., construction zones), with semantic labels for point cloud segmentation in autonomous driving.

In addition to the public dataset, we also conduct experiments in a real-scene dataset, which is the self-acquisition point cloud data collected from the real oil field scene by the air-ground fusion acquisition equipment consisting of 16-line radar, camera, UAV, IMU, etc. The point cloud involves a real space covering about 30,000 square metres, and contains more open weakly textured areas in the point cloud scene, as well as fewer samples of elongated objects in the categories of pipelines, fences, poles, road surfaces, containers, scaffolding, pumping machines, etc. In addition, we experimentally test the performance of the proposed method on part segmentation tasks on the ShapeNet \citep{23-shapenet} dataset. ShapeNet-Part \citep{23-shapenet} focuses on part segmentation with 16,881 models from 16 categories annotated with 50 part classes (e.g., airplane wings, chair legs), making it a benchmark for 3D point cloud part segmentation tasks.

\begin{figure*}[t]
\centering
\includegraphics[width=0.9\textwidth]{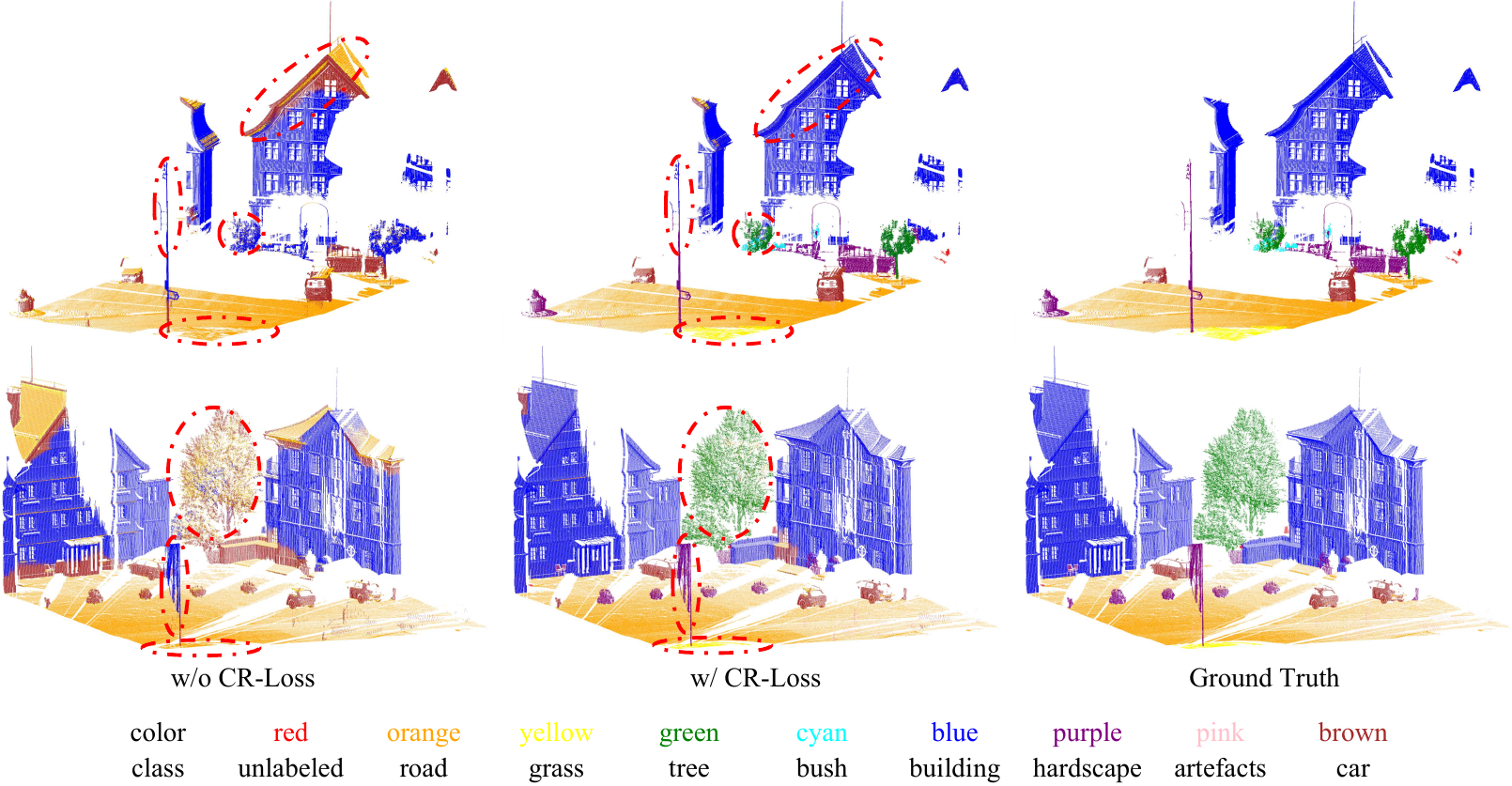}
\caption{Qualitative experiment and ablation experiment results of semantic segmentation of outdoor point cloud dataset Semantic3D \citep{22-semantic3d}.}
\label{fig5}
\end{figure*}

\begin{table}[!t]
\caption{Quantitative comparison experiment results of semantic segmentation of indoor point cloud dataset S3DIS \citep{21-s3dis} (6-fold validation).}
\label{tab2}
\centering
{\small
\begin{tabular}{l|ccc}
\toprule
  \makebox[12mm][l]{Method} & \makebox[10mm]{OA} & \makebox[10mm]{mAcc} & \makebox[10mm]{mIoU}\\\hline
  PointNet \citep{8-pointnet} & 78.6 & 66.2 & 47.6\\
  PointNet++ \citep{9-pointnet++} & 81.0 & 67.1 & 54.5\\
  KPConv \citep{7-kpconv} & - & 79.1 & 70.6\\
  RandLA-Net \citep{11-randlanet} & 88.0 & \underline{82.0} & 70.0\\
  PT \citep{14-pt} & \underline{90.2} & 81.9 & 73.5\\
  SuperCluster \citep{20-supercluster} & - & - & \underline{75.3}\\\hline
  Ours & \textbf{91.7} & \textbf{85.0} & \textbf{75.8}\\
\bottomrule
\end{tabular}}
\end{table}

\subsubsection{Experimental Details}
Following the practice of new point cloud segmentation models in recent years \citep{18-superpt,19-spotr,20-supercluster}, we use Overall Accuracy (OA), mean Accuracy (mAcc), mean Intersection over Union (mIoU) for semantic segmentation and Class (Cls.) mIoU, Instance (Ins.) mIoU for part segmentation. The deployment platforms for the experiments are all PyTorch, the computing platform for the indoor scene experiments is a server equipped with two Titan RTX (24G*2 video memory), and the computing platform for the outdoor scene experiments is a host computer equipped with an RTX4090 (24G video memory). The optimiser in the training process is selected as Adam, and the initial learning rate is set to 0.006. For the hyperparameter settings involved in the article, the number of discrete parts of the densities is generally set to 5 by default in the indoor scenario, while the number of discrete parts is set to 7 in the outdoor scenario. And the hyperparameter used in the loss function to balance the two loss terms is set to 0.5 by default, in order to prevent either loss component from dominating, while optimally exploiting their complementary strengths in dealing with class-imbalanced distributions and spatial consistency constraints. Experiments on the ablation of the two hyperparameters mentioned above will be shown in the Section 4.5. In addition, the number of iterations of the proposed method is set to 100 by default for experimental training.

\subsection{Indoor Scene Experiment}

\begin{table}[!t]
\caption{Quantitative comparison experiment results of semantic segmentation of indoor point cloud dataset ScanNet \citep{33-scannet}.}
\label{tab3}
\centering
{\small
\begin{tabular}{l|c}
\toprule
  \makebox[12mm][l]{Method} & \makebox[15mm]{mIoU}\\\hline
  PointNet++ \citep{9-pointnet++} & 53.5\\
  KPConv \citep{7-kpconv} & 69.2\\
  ST \citep{16-st} & 74.3\\
  PTv2 \citep{17-ptv2} & 75.4\\
  OctFormer \citep{34-octformer} & \underline{75.7}\\\hline
  Ours & \textbf{77.5}\\
\bottomrule
\end{tabular}}
\end{table}


Table \ref{tab1} summarises the results for the indoor scene dataset S3DIS \citep{21-s3dis} Area 5 experimental scenarios. The experimental results are listed as overall and category-level accuracy metrics for the comparison methods \citep{13-pct,16-st,8-pointnet,7-kpconv,18-superpt,19-spotr} and for the proposed method. \textbf{Bolded} data represent the best results for the corresponding metrics, while \underline{underlining} represents the second best results for the metrics. In the overall indexes, they are better than other comparative methods and state-of-the-art methods, and the proposed method category mIoU are also the best in six of the 13 categories, while the other seven categories have also reached the sub-optimal level. Table \ref{tab2} summarises the results under the six-fold cross-validation experimental scenario for the indoor scene dataset S3DIS \citep{21-s3dis}. The overall accuracy, average accuracy, and average cross-validation ratio values of the comparison method \citep{14-pt,8-pointnet,9-pointnet++,11-randlanet,7-kpconv,20-supercluster}, as well as the method of this paper, are listed in the experimental results. In addition, the experimental results on the validation set of the ScanNet \citep{33-scannet} dataset in Table \ref{tab3} shows that our method have a great improvement among other methods.

\begin{figure*}[t]
\centering
\includegraphics[width=0.9\textwidth]{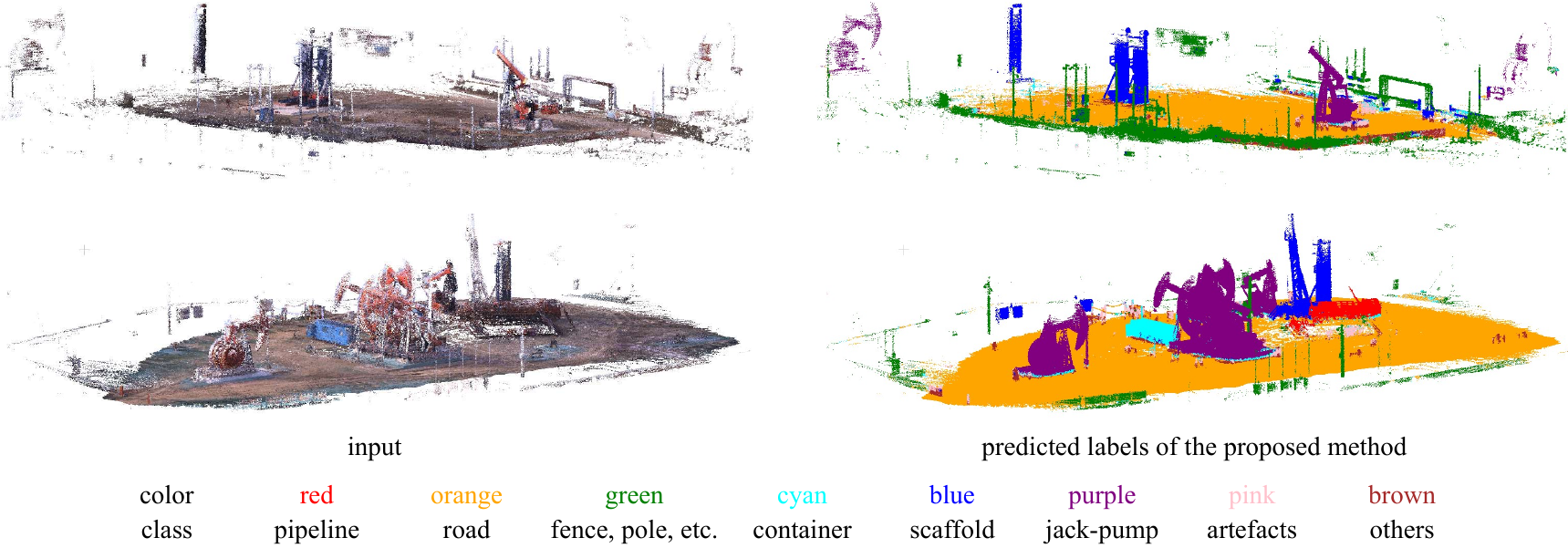}
\vspace{-0.2cm}
\caption{Qualitative segmentation results of the proposed method on self-collected point cloud data of actual oil field scenes.}
\label{fig6}
\end{figure*}

\begin{table}[t]
\caption{Quantitative comparison experiment results of semantic segmentation of outdoor point cloud dataset SemanticKITTI \citep{38-semantickitti} on the validation set. $\ddag$ represents the results of testing using publicly available pre-trained models.}
\label{tab5}
\centering
{\small
\begin{tabular}{l|cc}
\toprule
  \makebox[18mm][l]{Method} & \makebox[12mm]{mIoU} & \makebox[12mm]{OA $\ddag$}\\\hline
  SphereFormer\citep{39-sphereformer} & 67.8 & 91.5\\
  WaffleIron\citep{40-waffleiron} & 68.0 & -\\
  2DPASS\citep{41-2dpass} & 69.3 & -\\
  PTv2\citep{17-ptv2} & 70.3 & 91.9\\
  OA-CNNs\citep{42-oacnns} & 70.6 & 92.2\\
  PPT+SparseUNet\citep{43-ppt} & 71.4 & 91.6\\
  PTv3+PPT\citep{43-ppt,44-ptv3} & \underline{72.3} & \underline{92.5}\\\hline
  Ours & \textbf{72.9} & \textbf{92.6}\\
\bottomrule
\end{tabular}}
\end{table}

\begin{table}[t]
\caption{Quantitative comparison experiment results of semantic segmentation of outdoor point cloud dataset nuScenes \citep{45-nuscenes}.}
\label{tab6}
\centering
{\small
\begin{tabular}{l|c}
\toprule
  \makebox[18mm][l]{Method} & \makebox[15mm]{mIoU}\\\hline
  OA-CNNs\citep{42-oacnns} & 78.9\\
  FRNet\citep{46-frnet} & 79.0\\
  WaffleIron\citep{40-waffleiron} & 79.1\\
  SphereFormer\citep{39-sphereformer} & 79.5\\
  PTv2\citep{17-ptv2} & 80.2\\
  PPT+SparseUNet\citep{43-ppt} & 78.6\\
  PTv3\citep{43-ppt,44-ptv3} & \underline{80.4}\\\hline
  Ours & \textbf{80.6}\\
\bottomrule
\end{tabular}}
\end{table}

Compared to the MLP-based as well as the point convolution methods, the proposed method, as an attention-based method, the 3D points input to the network can have a larger effective sensory field, which makes its segmentation accuracy significantly better than these comparative methods. Compared to other methods, our proposed DbGLA module determines different sizes of local window sizes to compute the local self-attention for different local areas within the point cloud densities, so that more dense parts of the point cloud can be allocated more windows to compute the local attention, resulting in a higher segmentation accuracy for the categories of windows, chairs, sofas, bookcases, etc., which have a smaller sample of objects compared to the ceiling, floor, and so on. Our method obtains contextual information at the level of the whole scene by computing the global attention between local areas to enhance the long range modelling capability of the network, giving the proposed backbone network a stronger feature extraction capability and thus a higher overall segmentation accuracy.

\begin{table}[t]
\caption{Quantitative comparison experiment results of semantic segmentation of outdoor point cloud dataset Waymo \citep{47-waymo}.}
\label{tab7}
\centering
{\small
\begin{tabular}{l|cc}
\toprule
  \makebox[18mm][l]{Method} & \makebox[15mm]{mIoU} & \makebox[15mm]{mAcc}\\\hline
  MinkUNet \citep{48-minkunet} & 65.9 & 76.6\\
  SPVNAS \citep{49-spvnas} & 67.4 & -\\
  Cylinder3D \citep{50-cylinder} & 66.0 & -\\
  SphereFormer\citep{39-sphereformer} & 69.9 & -\\
  PTv2\citep{17-ptv2} & 70.6 & 80.2\\
  PTv3\citep{44-ptv3} & \underline{71.3} & \underline{80.5}\\\hline
  Ours & \textbf{71.7} & \textbf{80.9}\\
\bottomrule
\end{tabular}}
\end{table}

\subsection{Outdoor Scene Experiment}
Experiments on the SemanticKITTI validation set (Table \ref{tab5}) show that our method has an accuracy improvement of at least 0.6$\%$ in mIoU compared to existing methods including SOTA Transformer method PTV3 \citep{44-ptv3}. Experiments on the nuScenes datasets (Table \ref{tab6}) also show that proposed method outperforms the methods including training on other datasets (PPT+SparseUNet \citep{43-ppt}) for mIoU. We show the performance of our method and SOTA methods on Waymo \citep{47-waymo} in Table \ref{tab7}. In the experiments on this dataset, our method also ranks first among methods that do not use other datasets, with an accuracy advantage of 0.4$\%$ to 5.8$\%$ over other methods that also experiment on this dataset. This shows that our method has superior segmentation accuracies on several outdoor datasets.

The results in Figure \ref{fig5} represent the results of qualitative comparison experiments in employing two partial scenes from the Semantic3D \citep{22-semantic3d} validation set of outdoor scene datasets. The left column of the figure shows the segmentation results of this paper's method without incorporating the semantic coding loss function CR-Loss, the middle column shows the segmentation results after incorporating both CR-Loss as well as the DbGLA module, and the right column shows the data truth values of the corresponding scenes. 

It can be seen that the proposed method is more effective in segmenting outdoor scenes. Compared with the method without adding the proposed 3D CR-Loss, the method of adding the 3D CR-Loss based on the existing DbGLA module can balance the network's attention among different categories and sizes of objects. As a result, in the experimental results, as shown in the categories with fewer samples such as grass, hardscapes, trees, shrubs, etc., which are marked with red circle in Figure \ref{fig5}, the segmentation effect of our method is significantly improved in the small-sample categories.

In addition to experiments on public dataset, we collect point clouds in real scenes for additional segmentation experiments. As shown in Figure \ref{fig6}, it is the qualitative result demonstration of the proposed method for segmentation on point cloud data of multiple self-produced oil field scenes after pre-training in the outdoor public dataset. Through the visual demonstration, it can be seen that the proposed method can better achieve more equal attention to the large sample categories and small sample categories, and large objects and small objects when processing point clouds. Especially for the long and thin objects such as poles, railings, pipes, which are more distributed in the actual scene of the oil field, they can be identified and processed better.

\begin{table}[!t]
\caption{Quantitative comparison experimental results of part segmentation of 3D point cloud dataset ShapeNet-Part \citep{23-shapenet}.}
\label{tab4}
\centering
{\small
\begin{tabular}{l|cc}
\toprule
  \makebox[14mm][l]{Method} & \makebox[15mm]{Cls. mIoU} & \makebox[15mm]{Ins. mIoU}\\\hline
  PointNet \citep{8-pointnet} & 80.4 & 83.7\\
  PointNet++ \citep{9-pointnet++} & 81.9 & 85.1\\
  KPConv \citep{7-kpconv} & 85.0 & 86.2\\
  PCT \citep{13-pct} & - & 86.4\\
  PT \citep{14-pt} & 83.7 & \underline{86.6}\\
  ST \citep{16-st} & \underline{85.1} & 86.6\\\hline
  Ours & \textbf{85.2} & \textbf{87.0}\\
\bottomrule
\end{tabular}}
\end{table}

\begin{table}[t]
\caption{Ablation studies on indoor point cloud dataset S3DIS \citep{21-s3dis} Area 5. SLA stands for shifted-window local attention. VGA represent voxel global attention. WCE means weighted cross entropy loss. $\Delta$ represents the improvement in mIoU compared to the method using only shifted-window local attention and weighted cross-entropy loss.}
\label{tab8}
\centering
{\small
\begin{tabular}{ll|c|c}
\toprule
  \makebox[12mm][l]{Attention} & \makebox[12mm][l]{Loss Term} & \makebox[12mm]{mIoU} & \makebox[8mm]{$\Delta$}\\\hline
  SLA & WCE & 70.6 & +0.0\\
  SLA & WCE and ours CR-Loss & 71.0 & +0.4\\
  SLA and VGA & WCE & 71.2 & +0.6\\
  SLA and VGA & WCE and ours CR-Loss & 71.5 & +0.9\\
  Ours DbGLA & WCE & 71.7 & +1.1\\
  Ours DbGLA & WCE and ours CR-Loss & 72.5 & +1.9\\
\bottomrule
\end{tabular}}
\end{table}

\subsection{Part Segmentation Experiment}
Table \ref{tab4} shows the experimental results for the 3D point cloud part segmentation task dataset ShapeNet \citep{23-shapenet}, where the comparison methods \citep{13-pct,14-pt,16-st,8-pointnet,9-pointnet++,7-kpconv}, and the Class mIoU and Instance mIoU of the proposed method are listed respectively. From this comparison experiment, it can be seen that the proposed method outperforms the other compared methods in a variety of metrics in the part segmentation task, again thanks to the newly proposed DbGLA attention module. For local areas with different densities, our method uses windows of different sizes in order to compute the local attention within the corresponding local area, and with the addition of the proposed 3D CR-Loss, it further balances the network's attention to the different parts of each object, thus improving the segmentation accuracy in the part segmentation task.

\begin{table}[t]
\caption{Ablation studies on outdoor point cloud dataset SemanticKITTI \citep{38-semantickitti} validation set. $\Delta$ represents the improvement in mIoU compared to the method using only shifted-window local attention and weighted cross-entropy loss.}
\label{tab9}
\centering
{\small
\begin{tabular}{ll|c|c}
\toprule
  \makebox[12mm][l]{Attention} & \makebox[12mm][l]{Loss Term} & \makebox[12mm]{mIoU} & \makebox[8mm]{$\Delta$}\\\hline
  SLA & WCE & 69.7 & +0.0\\
  SLA & WCE and ours CR-Loss & 70.6 & +0.9\\
  SLA and VGA & WCE & 71.4 & +1.7\\
  SLA and VGA & WCE and ours CR-Loss & 71.9 & +2.2\\
  Ours DbGLA & WCE & 72.2 & +2.5\\
  Ours DbGLA & WCE and ours CR-Loss & 72.9 & +3.2\\
\bottomrule
\end{tabular}}
\end{table}

\begin{table}[t]
\caption{Ablation studies of hyperparameters on S3DIS \citep{21-s3dis} Area 5 and SemanticKITTI \citep{38-semantickitti} validation set.}
\label{tab10}
\centering
{\small
\begin{tabular}{l|ccccc}
\toprule
  \makebox[12mm][l]{Value of $\lambda$} & \makebox[5mm][l]{0.1} & \makebox[5mm][l]{0.3} & \makebox[5mm][l]{0.5} & \makebox[5mm][l]{0.7} & \makebox[5mm][l]{0.9}\\\hline
  S3DIS mIoU & 71.3 & 71.9 & \textbf{72.5} & 72.2 & 71.8\\
  SemanticKITTI mIoU & 71.8 & 72.4 & \textbf{72.9} & 72.5 & 72.3\\\hline
  Num. of discrete parts & 1 & 3 & 5 & 7 & 9\\\hline
  S3DIS mIoU & 71.0 & 71.7 & \textbf{72.5} & 72.3 & 72.1\\
  SemanticKITTI mIoU & 70.6 & 71.5 & 72.3 & \textbf{72.9} & 72.8\\
\bottomrule
\end{tabular}}
\end{table}

\begin{table}[t]
\caption{Ablation studies of the number of iterations on S3DIS \citep{21-s3dis} Area 5 and SemanticKITTI \citep{38-semantickitti} validation set.}
\label{tab11}
\centering
{\small
\begin{tabular}{l|ccccc}
\toprule
  \makebox[12mm][l]{Number of iterations} & \makebox[5mm][l]{20} & \makebox[5mm][l]{40} & \makebox[5mm][l]{60} & \makebox[5mm][l]{80} & \makebox[5mm][l]{100}\\\hline
  S3DIS mIoU & 35.2 & 52.8 & 65.1 & 70.9 & \textbf{72.5}\\
  SemanticKITTI mIoU & 18.9 & 48.3 & 65.5 & 70.3 & \textbf{72.9}\\\hline
  Number of iterations & 110 & 120 & 130 & 140 & 150\\\hline
  S3DIS mIoU & 72.4 & 72.3 & 72.2 & 72.3 & 72.1\\
  SemanticKITTI mIoU & 72.8 & 72.5 & 72.7 & 72.6 & 72.5\\
\bottomrule
\end{tabular}}
\end{table}

\subsection{Ablation Experiment}
Tables \ref{tab8} and \ref{tab9} give the results of the ablation experiments comparing our proposed DbGLA attention mechanism with the commonly used shifted-window local attention and combined voxelised global attention on the indoor and outdoor datasets, respectively, and show the ablation experiments comparing the proposed CR-Loss with the commonly used loss. The experimental results demonstrate that the proposed DbGLA attention mechanism has an accuracy improvement of 0.5$\%$$\sim$1.0$\%$ over the existing attention mechanisms on both indoor and outdoor datasets. The proposed CR-Loss also allows for a further improvement in mIoU of 0.3$\%$$\sim$0.9$\%$.

In addition, Table \ref{tab10} shows the ablation studies of the two hyperparameters involved in the proposed method. For $\lambda$ which is used to balance the two loss terms, we set $\lambda = \{0.1, 0.3, 0.5, 0.7, 0.9\}$ for the experimental group, and it can be seen that the network accuracy reaches its peak when $\lambda = 0.5$, so we set the value of this hyperparameter $\lambda$ to $0.5$ by default in our experiments. For the other hyperparameter, the number of discrete parts of the point cloud densities, we set $\{1, 3, 5, 7, 9\}$ for the experimental groups. As can be seen from the data in the table, when the value of discrete parts is set to $1$, the network accuracy is the lowest among all the experimental groups, proving the effectiveness of our proposed density-based attention mechanism. Moreover, for the indoor dataset and the outdoor dataset, the network accuracy comes to the peak when the number of discrete parts is set to $5$ and $7$, respectively, and thus we conduct the experiments according to the corresponding hyperparameter settings.

The ablation study on the number of iterations, presented in Table \ref{tab11}, clearly demonstrates that 100 is the optimal value. As shown in the table, the mIoU on both the S3DIS and SemanticKITTI datasets increases steadily with the number of iterations, peaking at 72.5 and 72.9, respectively, at 100 iterations. Crucially, beyond this point, further training yields no improvement; instead, the performance plateaus and even slightly decreases, indicating a signs of overfitting. Therefore, setting the number of iterations to 100 achieves the best performance while ensuring training efficiency.

Furthermore, as shown in Table \ref{tab1} and Figure \ref{fig5}, the effectiveness of the DbGLA attention mechanism proposed in this paper as well as the proposed loss function CR-Loss in the point cloud segmentation task is verified by performing quantitative ablation experiments in the indoor scene dataset S3DIS \citep{21-s3dis} and qualitative ablation experiments in the outdoor scene dataset Semantic3D \citep{22-semantic3d}. Compared with the network that incorporates both the DbGLA attention mechanism and the CR-Loss, the network that incorporates only CR-Loss does not apply a stronger degree of data extraction and attention to denser areas, and thus does not differentiate the treatment of local areas of point clouds with different degrees of densities, resulting in a decrease in the overall accuracy of 0.5$\%$$\sim$2.2$\%$, and a decrease in the mIoU of some categories of 2.1$\%$$\sim$8.6$\%$, reflecting the fact that the DbGLA attention mechanism and the proposed loss function have been used for the segmentation of point clouds. Compared to the network with only the DbGLA attention mechanism and the additional introduction of CR-Loss, there is a lack of equal attention to large and small targets, and some small objects such as smaller man-made landscapes and low bushes do not get the correct segmentation results, reflecting the fact that the newly added category response loss function CR-Loss in this paper is favourable for semantic segmentation of smaller samples as well as samples of small categories.

\section{Conclusions}
In this paper, we propose a point cloud segmentation method that fuses density-aware local attention with global attention to build a balanced perception of local details and scene context. Using the proposed Category Response Loss, our method achieves a better balance among different categories and sizes of objects, so that it can better discriminate the harder-to-recognise small objects. Quantitative as well as qualitative experimental results on multiple publicly available datasets demonstrate the effectiveness of the proposed attention mechanism and the novel category response loss. It should be noted that the effectiveness of the method can be influenced by the density and quality of the input point clouds. Furthermore, while the current framework is designed for indoor scenes, its extension to extremely large-scale outdoor environments presents computational and architectural challenges that warrant further investigation. The follow-up work can focus on analysing the shortcomings and corresponding improvement methods on researching the existing Transformer-based 3D semantic segmentation network for indoor scenes to the 3D segmentation task of outdoor scenes.

\section*{CRediT authorship contribution statement}
Chade Li: Writing - review $\&$ editing, Writing - original draft, Visualization, Validation, Resources, Methodology, Investigation, Formal analysis. Pengju Zhang: Data curation. Jiaming Zhang: Writing - review $\&$ editing. Yihong Wu: Supervision.

\section*{Acknowledgment}
This work was supported by the National Natural Science Foundation of China under Grant No. 62403459, and Beijing Natural Science Foundation under Grant No. L241012.

\bibliographystyle{elsarticle-num}
\bibliography{ivc}

\begin{thebibliography}{10}
\expandafter\ifx\csname url\endcsname\relax
  \def\url#1{\texttt{#1}}\fi
\expandafter\ifx\csname urlprefix\endcsname\relax\def\urlprefix{URL }\fi
\expandafter\ifx\csname href\endcsname\relax
  \def\href#1#2{#2} \def\path#1{#1}\fi

\bibitem{12-attention}
A.~Vaswani, N.~Shazeer, N.~Parmar, J.~Uszkoreit, L.~Jones, A.~N. Gomez, L.~Kaiser, I.~Polosukhin, Attention is all you need, CoRR abs/1706.03762 (2017).
\newblock \href {http://arxiv.org/abs/1706.03762} {\path{arXiv:1706.03762}}.

\bibitem{13-pct}
M.-H. Guo, J.-X. Cai, Z.-N. Liu, T.-J. Mu, R.~R. Martin, S.-M. Hu, Pct: Point cloud transformer, Computational Visual Media 7 (2021) 187--199.

\bibitem{36-3crossnet}
X.-F. Han, Z.-Y. He, J.~Chen, G.-Q. Xiao, 3crossnet: Cross-level cross-scale cross-attention network for point cloud representation, IEEE Robotics and Automation Letters 7~(2) (2022) 3718--3725.

\bibitem{37-pointbert}
X.~Yu, L.~Tang, Y.~Rao, T.~Huang, J.~Zhou, J.~Lu, Point-bert: Pre-training 3d point cloud transformers with masked point modeling, in: Proceedings of the IEEE/CVF conference on computer vision and pattern recognition, 2022, pp. 19313--19322.

\bibitem{14-pt}
H.~Zhao, L.~Jiang, J.~Jia, P.~H. Torr, V.~Koltun, Point transformer, in: Proceedings of the IEEE/CVF international conference on computer vision, 2021, pp. 16259--16268.

\bibitem{16-st}
X.~Lai, J.~Liu, L.~Jiang, L.~Wang, H.~Zhao, S.~Liu, X.~Qi, J.~Jia, Stratified transformer for 3d point cloud segmentation, in: Proceedings of the IEEE/CVF conference on computer vision and pattern recognition, 2022, pp. 8500--8509.

\bibitem{35-pptnet}
L.~Hui, H.~Yang, M.~Cheng, J.~Xie, J.~Yang, Pyramid point cloud transformer for large-scale place recognition, in: Proceedings of the IEEE/CVF International Conference on Computer Vision, 2021, pp. 6098--6107.

\bibitem{4-subsc}
B.~Graham, M.~Engelcke, L.~Van Der~Maaten, 3d semantic segmentation with submanifold sparse convolutional networks, in: Proceedings of the IEEE conference on computer vision and pattern recognition, 2018, pp. 9224--9232.

\bibitem{5-4dstc}
C.~Choy, J.~Gwak, S.~Savarese, 4d spatio-temporal convnets: Minkowski convolutional neural networks, in: Proceedings of the IEEE/CVF conference on computer vision and pattern recognition, 2019, pp. 3075--3084.

\bibitem{1-deep3ds}
F.~J. Lawin, M.~Danelljan, P.~Tosteberg, G.~Bhat, F.~S. Khan, M.~Felsberg, Deep projective 3d semantic segmentation, in: Computer Analysis of Images and Patterns: 17th International Conference, CAIP 2017, Ystad, Sweden, August 22-24, 2017, Proceedings, Part I 17, Springer, 2017, pp. 95--107.

\bibitem{2-squeezeseg}
B.~Wu, A.~Wan, X.~Yue, K.~Keutzer, Squeezeseg: Convolutional neural nets with recurrent crf for real-time road-object segmentation from 3d lidar point cloud, in: 2018 IEEE international conference on robotics and automation (ICRA), IEEE, 2018, pp. 1887--1893.

\bibitem{3-polarnet}
Y.~Zhang, Z.~Zhou, P.~David, X.~Yue, Z.~Xi, B.~Gong, H.~Foroosh, Polarnet: An improved grid representation for online lidar point clouds semantic segmentation, in: Proceedings of the IEEE/CVF conference on computer vision and pattern recognition, 2020, pp. 9601--9610.

\bibitem{8-pointnet}
C.~R. Qi, H.~Su, K.~Mo, L.~J. Guibas, Pointnet: Deep learning on point sets for 3d classification and segmentation, in: Proceedings of the IEEE conference on computer vision and pattern recognition, 2017, pp. 652--660.

\bibitem{9-pointnet++}
C.~R. Qi, L.~Yi, H.~Su, L.~J. Guibas, Pointnet++: Deep hierarchical feature learning on point sets in a metric space, Advances in neural information processing systems 30 (2017).

\bibitem{10-deepgcns}
G.~Li, M.~Muller, A.~Thabet, B.~Ghanem, Deepgcns: Can gcns go as deep as cnns?, in: Proceedings of the IEEE/CVF international conference on computer vision, 2019, pp. 9267--9276.

\bibitem{11-randlanet}
Q.~Hu, B.~Yang, L.~Xie, S.~Rosa, Y.~Guo, Z.~Wang, N.~Trigoni, A.~Markham, Learning semantic segmentation of large-scale point clouds with random sampling, IEEE Transactions on Pattern Analysis and Machine Intelligence 44~(11) (2021) 8338--8354.

\bibitem{28-thickseg}
Q.~Gao, X.~Shen, Thickseg: Efficient semantic segmentation of large-scale 3d point clouds using multi-layer projection, Image and Vision Computing 108 (2021) 104161.

\bibitem{6-pointcnn}
Y.~Li, R.~Bu, M.~Sun, W.~Wu, X.~Di, B.~Chen, Pointcnn: Convolution on x-transformed points, Advances in neural information processing systems 31 (2018).

\bibitem{7-kpconv}
H.~Thomas, C.~R. Qi, J.-E. Deschaud, B.~Marcotegui, F.~Goulette, L.~J. Guibas, Kpconv: Flexible and deformable convolution for point clouds, in: Proceedings of the IEEE/CVF international conference on computer vision, 2019, pp. 6411--6420.

\bibitem{26-stpca}
J.~Lin, K.~Zhong, T.~Gong, X.~Zhang, N.~Wang, Point cloud segmentation neural network with same-type point cloud assistance, Image and Vision Computing 152 (2024) 105331.

\bibitem{30-hcfs3d}
J.~Tan, K.~Wang, L.~Chen, G.~Zhang, J.~Li, X.~Zhang, Hcfs3d: Hierarchical coupled feature selection network for 3d semantic and instance segmentation, Image and Vision Computing 109 (2021) 104129.

\bibitem{29-mdcgcns}
W.~Tang, G.~Qiu, Dense graph convolutional neural networks on 3d meshes for 3d object segmentation and classification, Image and Vision Computing 114 (2021) 104265.

\bibitem{31-dgcnn}
Y.~Wang, Y.~Sun, Z.~Liu, S.~E. Sarma, M.~M. Bronstein, J.~M. Solomon, Dynamic graph cnn for learning on point clouds, ACM Trans. Graph. 38~(5) (Oct. 2019).

\bibitem{24-3dmv}
A.~Dai, M.~Nie{\ss}ner, 3dmv: Joint 3d-multi-view prediction for 3d semantic scene segmentation, in: Proceedings of the European Conference on Computer Vision (ECCV), 2018, pp. 452--468.

\bibitem{25-xmuda}
M.~Jaritz, T.-H. Vu, R.~d. Charette, E.~Wirbel, P.~P{\'e}rez, xmuda: Cross-modal unsupervised domain adaptation for 3d semantic segmentation, in: Proceedings of the IEEE/CVF conference on computer vision and pattern recognition, 2020, pp. 12605--12614.

\bibitem{15-fpt}
C.~Park, Y.~Jeong, M.~Cho, J.~Park, Fast point transformer, in: Proceedings of the IEEE/CVF conference on computer vision and pattern recognition, 2022, pp. 16949--16958.

\bibitem{17-ptv2}
X.~Wu, Y.~Lao, L.~Jiang, X.~Liu, H.~Zhao, Point transformer v2: Grouped vector attention and partition-based pooling, Advances in Neural Information Processing Systems 35 (2022) 33330--33342.

\bibitem{18-superpt}
D.~Robert, H.~Raguet, L.~Landrieu, Efficient 3d semantic segmentation with superpoint transformer, in: Proceedings of the IEEE/CVF International Conference on Computer Vision, 2023, pp. 17195--17204.

\bibitem{19-spotr}
J.~Park, S.~Lee, S.~Kim, Y.~Xiong, H.~J. Kim, Self-positioning point-based transformer for point cloud understanding, in: Proceedings of the IEEE/CVF conference on computer vision and pattern recognition, 2023, pp. 21814--21823.

\bibitem{20-supercluster}
D.~Robert, H.~Raguet, L.~Landrieu, Scalable 3d panoptic segmentation as superpoint graph clustering, in: 2024 International Conference on 3D Vision (3DV), IEEE, 2024, pp. 179--189.

\bibitem{27-iamtm}
T.~Yuan, Y.~Yu, X.~Wang, Semantic segmentation of large-scale point clouds by integrating attention mechanisms and transformer models, Image and Vision Computing 146 (2024) 105019.

\bibitem{38-semantickitti}
J.~Behley, M.~Garbade, A.~Milioto, J.~Quenzel, S.~Behnke, C.~Stachniss, J.~Gall, Semantickitti: A dataset for semantic scene understanding of lidar sequences, in: Proceedings of the IEEE/CVF International Conference on Computer Vision (ICCV), 2019, pp. 9296--9306.

\bibitem{32-dbscan}
M.~Ester, H.-P. Kriegel, J.~Sander, X.~Xu, A density-based algorithm for discovering clusters in large spatial databases with noise, in: Proceedings of the Second International Conference on Knowledge Discovery and Data Mining, KDD'96, AAAI Press, 1996, p. 226–231.

\bibitem{21-s3dis}
I.~Armeni, O.~Sener, A.~R. Zamir, H.~Jiang, I.~Brilakis, M.~Fischer, S.~Savarese, 3d semantic parsing of large-scale indoor spaces, in: Proceedings of the IEEE conference on computer vision and pattern recognition, 2016, pp. 1534--1543.

\bibitem{42-oacnns}
B.~Peng, X.~Wu, L.~Jiang, Y.~Chen, H.~Zhao, Z.~Tian, J.~Jia, Oa-cnns: Omni-adaptive sparse cnns for 3d semantic segmentation, in: Proceedings of the IEEE/CVF Conference on Computer Vision and Pattern Recognition (CVPR), 2024, pp. 21305--21315.

\bibitem{48-minkunet}
C.~Choy, J.~Gwak, S.~Savarese, 4d spatio-temporal convnets: Minkowski convolutional neural networks, in: 2019 IEEE/CVF Conference on Computer Vision and Pattern Recognition (CVPR), 2019, pp. 3070--3079.

\bibitem{49-spvnas}
H.~Tang, Z.~Liu, S.~Zhao, Y.~Lin, J.~Lin, H.~Wang, S.~Han, Searching efficient 3d architectures with sparse point-voxel convolution, in: A.~Vedaldi, H.~Bischof, T.~Brox, J.-M. Frahm (Eds.), Computer Vision -- ECCV 2020, Springer International Publishing, Cham, 2020, pp. 685--702.

\bibitem{50-cylinder}
X.~Zhu, H.~Zhou, T.~Wang, F.~Hong, Y.~Ma, W.~Li, H.~Li, D.~Lin, Cylindrical and asymmetrical 3d convolution networks for lidar segmentation, in: 2021 IEEE/CVF Conference on Computer Vision and Pattern Recognition (CVPR), 2021, pp. 9934--9943.

\bibitem{40-waffleiron}
G.~Puy, A.~Boulch, R.~Marlet, Using a waffle iron for automotive point cloud semantic segmentation, in: 2023 IEEE/CVF International Conference on Computer Vision (ICCV), 2023, pp. 3356--3366.

\bibitem{41-2dpass}
X.~Yan, J.~Gao, C.~Zheng, C.~Zheng, R.~Zhang, S.~Cui, Z.~Li, 2dpass: 2d priors assisted semantic segmentation on lidar point clouds, in: S.~Avidan, G.~Brostow, M.~Ciss{\'e}, G.~M. Farinella, T.~Hassner (Eds.), Computer Vision -- ECCV 2022, Springer Nature Switzerland, Cham, 2022, pp. 677--695.

\bibitem{46-frnet}
X.~Xu, L.~Kong, H.~Shuai, Q.~Liu, Frnet: Frustum-range networks for scalable lidar segmentation (2024).
\newblock \href {http://arxiv.org/abs/2312.04484} {\path{arXiv:2312.04484}}.

\bibitem{34-octformer}
P.-S. Wang, Octformer: Octree-based transformers for 3d point clouds, ACM Transactions on Graphics (TOG) 42~(4) (2023) 1--11.

\bibitem{39-sphereformer}
X.~Lai, Y.~Chen, F.~Lu, J.~Liu, J.~Jia, Spherical transformer for lidar-based 3d recognition, in: Proceedings of the IEEE/CVF Conference on Computer Vision and Pattern Recognition (CVPR), 2023, pp. 17545--17555.

\bibitem{44-ptv3}
X.~Wu, L.~Jiang, P.-S. Wang, Z.~Liu, X.~Liu, Y.~Qiao, W.~Ouyang, T.~He, H.~Zhao, Point transformer v3: Simpler faster stronger, in: Proceedings of the IEEE/CVF Conference on Computer Vision and Pattern Recognition (CVPR), 2024, pp. 4840--4851.

\bibitem{33-scannet}
A.~Dai, A.~X. Chang, M.~Savva, M.~Halber, T.~Funkhouser, M.~Nie{\ss}ner, Scannet: Richly-annotated 3d reconstructions of indoor scenes, in: Proceedings of the IEEE conference on computer vision and pattern recognition, 2017, pp. 5828--5839.

\bibitem{22-semantic3d}
T.~Hackel, N.~Savinov, L.~Ladicky, J.~D. Wegner, K.~Schindler, M.~Pollefeys, Semantic3d. net: A new large-scale point cloud classification benchmark, arXiv preprint arXiv:1704.03847 (2017).

\bibitem{45-nuscenes}
H.~Caesar, V.~Bankiti, A.~H. Lang, S.~Vora, V.~E. Liong, Q.~Xu, A.~Krishnan, Y.~Pan, G.~Baldan, O.~Beijbom, nuscenes: A multimodal dataset for autonomous driving, in: Proceedings of the IEEE/CVF Conference on Computer Vision and Pattern Recognition (CVPR), 2020, pp. 11621--11631.

\bibitem{47-waymo}
P.~Sun, H.~Kretzschmar, X.~Dotiwalla, A.~Chouard, V.~Patnaik, P.~Tsui, J.~Guo, Y.~Zhou, Y.~Chai, B.~Caine, V.~Vasudevan, W.~Han, J.~Ngiam, H.~Zhao, A.~Timofeev, S.~Ettinger, M.~Krivokon, A.~Gao, A.~Joshi, Y.~Zhang, J.~Shlens, Z.~Chen, D.~Anguelov, Scalability in perception for autonomous driving: Waymo open dataset, in: 2020 IEEE/CVF Conference on Computer Vision and Pattern Recognition (CVPR), 2020, pp. 2443--2451.

\bibitem{23-shapenet}
A.~X. Chang, T.~Funkhouser, L.~Guibas, P.~Hanrahan, Q.~Huang, Z.~Li, S.~Savarese, M.~Savva, S.~Song, H.~Su, et~al., Shapenet: An information-rich 3d model repository, arXiv preprint arXiv:1512.03012 (2015).

\bibitem{43-ppt}
X.~Wu, Z.~Tian, X.~Wen, B.~Peng, X.~Liu, K.~Yu, H.~Zhao, Towards large-scale 3d representation learning with multi-dataset point prompt training, in: Proceedings of the IEEE/CVF Conference on Computer Vision and Pattern Recognition (CVPR), 2024, pp. 19551--19562.

\end{thebibliography}

\end{document}